\documentclass[english]{IEEEtran}
\usepackage[T1]{fontenc}
\usepackage[latin9]{inputenc}
\usepackage{array}
\usepackage{multirow}
\usepackage{amsmath}
\usepackage{amssymb}
\usepackage{graphicx}

\makeatletter

\providecommand{\tabularnewline}{\\}

\usepackage{url}
\usepackage{hyperref}
\usepackage{cite}

\renewcommand\[{\begin{equation}}
\renewcommand\]{\end{equation}} 

\bstctlcite{IEEEexample:BSTcontrol}

\makeatother

\usepackage{babel}
\begin{document}
This work has been submitted to the IEEE for possible publication.
Copyright may be transferred without notice, after which this version
may no longer be accessible.

\pagebreak{}

\title{CT organ segmentation using GPU data augmentation, unsupervised labels
and IOU loss\thanks{Manuscript received 11/27/2018. The authors are with the Departments
of Electrical Engineering, Radiology and Biomedical Data Science at
Stanford University, Stanford, CA USA 94305 (email: blaine@stanford.edu,
darvinyi@stanford.edu, 19kaushiks@students.harker.org, tomomi@stanford.edu,
dlrubin@stanford.edu).}}

\author{Blaine Rister, Darvin Yi, Kaushik Shivakumar, Tomomi Nobashi and
Daniel L. Rubin\\
}
\maketitle
\begin{abstract}
Fully-convolutional neural networks have achieved superior performance
in a variety of image segmentation tasks. However, their training
requires laborious manual annotation of large datasets, as well as
acceleration by parallel processors with high-bandwidth memory, such
as GPUs. We show that simple models can achieve competitive accuracy
for organ segmentation on CT images when trained with extensive data
augmentation, which leverages existing graphics hardware to quickly
apply geometric and photometric transformations to 3D image data.
On 3 mm$^{3}$ CT volumes, our GPU implementation is 2.6-8$\times$faster
than a widely-used CPU version, including communication overhead.
We also show how to automatically generate training labels using rudimentary
morphological operations, which are efficiently computed by 3D Fourier
transforms. We combined fully-automatic labels for the lungs and bone
with semi-automatic ones for the liver, kidneys and bladder, to create
a dataset of 130 labeled CT scans. To achieve the best results from
data augmentation, our model uses the intersection-over-union (IOU)
loss function, a close relative of the Dice loss. We discuss its mathematical
properties and explain why it outperforms the usual weighted cross-entropy
loss for unbalanced segmentation tasks. We conclude that there is
no unique IOU loss function, as the naive one belongs to a broad family
of functions with the same essential properties. When combining data
augmentation with the IOU loss, our model achieves a Dice score of
78-92\% for each organ. The trained model, code and dataset will be
made publicly available, to further medical imaging research.
\end{abstract}


\section{Introduction}

Fully-convolutional neural networks (FCNs) have recently become the
de facto standard for medical image segmentation. These models simultaneously
detect and segment objects of interest from raw image data, eliminating
the need to design application-specific features. Current trends suggest
that FCN models are capable of performing a wide variety of organ
segmentation tasks, but they are limited by the scarcity of training
data and the computational issues inherent in volumetric image processing.
Additionally, medical images typically exhibit severe class imbalance,
for which the usual segmentation loss functions are poorly suited.
First, this paper addresses the data issue by creating a new dataset,
using manual annotation for some organs and unsupervised methods for
others, then performing 3D augmentations on this dataset using the
GPU. Next, the computational issues are addressed with a small, memory-efficient
model trained with a new loss function which excels at unbalanced
segmentation. The end result is a fast and robust model for multi-organ
segmentation. 

Typical FCNs contain millions of trained parameters across dozens
of layers. Training is almost always accelerated by graphics processing
units (GPUs), which provide massive parallelism and high memory bandwidth.
However, complex FCNs consume a prohibitive amount of high-bandwidth
memory when extended to process 3D images. This has led to a variety
of clever approaches for applying 2D models to 3D data. In contrast,
we argue that simpler 3D models can achieve competitive accuracy when
combined with extensive data augmentation. By randomly transforming
images at each training iteration, the network optimizes over a practically
infinite distribution of training images. We leverage GPU texture
sampling hardware to perform 3D data augmentation at negligible computational
cost. By combining geometric and photometric transformations, our
implementation minimizes memory usage and accelerates training. As
part of our data augmentation, we show how GPUs can quickly generate
random imaging noise. Our code is available as a small, easy-to-use
library\footnote{https://github.com/bbrister/cudaImageWarp} written
in Python, C++ and CUDA, which is 2.6-8$\times$faster than a SciPy
implementation, including communication overhead \cite{scipy:2001}.

Besides the input data, we found that the choice of loss function
is important for achieving high accuracy in unbalanced segmentation
tasks. In our experiments, data augmentation always decreases the
validation loss, but this only corresponds to an improvement in binary
segmentation accuracy for certain loss functions. Weighted cross-entropy,
the most common loss function for segmentation, performs poorly when
the classes are significantly unbalanced, as is often the case in
medical images. This is because it assigns a lesser penalty for false
positives than for false negatives. Instead, we propose the intersection-over-union
(IOU) loss function, which is closely related to the Dice loss, but
has the advantage of being a metric \cite{Spath:1981:JaccardMetric,Milletari:2016:Vnet}.
We analyze the mathematical properties of the IOU and Dice losses,
showing that they have approximately balanced penalties for each type
of error, and that each belongs to an infinite family of functions
having essentially the same properties. In our experiments, the IOU
loss significantly outperforms weighted cross-entropy, and results
are further improved by data augmentation.

Finally, we created a new dataset and applied these technologies to
CT organ segmentation, which is the task of detecting and delineating
organs in a CT scan.  Manual annotation of large datasets is impractical
for complex organs, such as the skeleton. To address this challenge,
we show how to automatically generate training labels using simple
image morphology, which is accelerated by 3D Fourier transforms. We
automatically generate labels for lungs and bone, which are then combined
with manual annotations for the liver, kidneys and bladder. Our experiments
suggest that the FCN is able to learn the ``average appearance''
of the organ, and is not distracted by intermittent failures in the
crude morphological algorithm. In addition to fully-automated organ
labels, we manually delineated the liver, kidneys and bladder in a
dataset of 130 CT scans from the Liver Tumor Segmentation Challenge
(LiTS), to create a free organ segmentation dataset with all five
organ classes \cite{Christ:2017:LiTs}. We then used this dataset
to train a multi-class FCN. Our model segments the lung, liver, bone,
kidneys and bladder with respective Dice scores of 91.8\%, 92.2\%,
79.3\%, 78.3\% and 83.7\%, consistent with results from the literature
which depend on more complex models \cite{Roth:2018:CascadedFCNOrganSeg,Gibson:2018:DenseVnet}.
We also evaluated the liver segmentation on the LiTS challenge test
set, with similar results. The trained model, code and dataset will
be made publicly available, to further medical imaging research.

Our main contributions, along with the structure of the paper, are
summarized as follows:
\begin{itemize}
\item Unsupervised algorithms for labeling the bones and lungs in CT scans
(section \ref{sec:Training-label-generation})
\item An efficient GPU implementation of 3D data augmentation (section \ref{sec:GPU-accelerated-data-augmentatio})
\item A new loss function which improves results for unbalanced segmentation
(section \ref{subsec:IOU-Loss})
\item A new dataset consisting of five organs labeled in 130 CT scans (section
\ref{subsec:Dataset})
\end{itemize}

\section{Related work}

This section describes the related work supporting each of our main
contributions\textendash first organ segmentation in general, then
using deep learning, then data augmentation, then weak supervision,
and finally loss functions. 

Organ segmentation has long been an active area of medical imaging
research. Approaches can be divided into two main categories: semi-automatic
and fully-automatic. The former group requires a user-generated starting
shape, which grows or shrinks into the organ of interest. These approaches
typically take into account intensity distributions, sharp edges and
even the shape of the target organ, and their success depends greatly
on the initialization. See Freiman et al.~for an example of semi-automatic
liver segmentation \cite{Freiman:2008:regionGrowingLiver}. 

The other category, fully-automatic methods, require no user input,
as they directly detect the object of interest in addition to delineating
its boundaries. These techniques can be divided into two main areas,
pattern recognition and atlas-based. Pattern recognition methods are
currently dominated by artificial neural networks such as FCNs, but
other other deep learning methods have been tried \cite{Shelhamer:2017:FCN}.
For example, Qin et al\@.~ used a CNN to classify super-pixels,
while Dong et al.~combined an FCN with a generative adversarial network
(GAN) \cite{Qin:2018:SuperpixelLiverSegmentation,Dong:2017:SiemensLiverSegGan}.
In contrast to pattern recognition, atlas-based methods work by warping,
or registering an image to a labeled atlas image. While atlas-based
methods can achieve high accuracy, inter-patient registration is computationally
expensive and extremely sensitive to changes in imaging conditions,
for example the absence of an organ. These methods are best suited
to stationary objects of consistent size and shape, such as the brain
\cite{Kalavathi:2016:SkullStrippingReview}. For whole-body imaging,
pattern recognition methods, and in particular deep learning, have
rapidly grown in popularity. For the sake of completeness, it is worth
noting that the distinction between these categories can be vague.
For example, Okada et al.~proposed a method which automatically generates
seeds for semi-automatic segmentation methods, based on the size and
location of a user-provided liver segmentation \cite{Okada:2015:Multi-organ}.

As in our work, many researchers are now applying FCNs to the task
of CT organ segmentation\cite{Roth:2018:CascadedFCNOrganSeg,Gibson:2018:DenseVnet,Milletari:2016:Vnet,Li:2018:H-DenseUNet_LiTS}.
However, most of these models are limited to a specific organ or body
region, around which they assume the images will be cropped. In addition,
some multi-organ projects are based on privately held data, and thus
can neither be replicated nor extended beyond the obvious limitations
of fine-tuning \cite{Roth:2018:CascadedFCNOrganSeg}. In contrast,
our simple model naturally applies to a wide variety of organs, and
requires no manual intervention in preparing the images. Leveraging
a memory-efficient model and resampling the images to a coarse $3$
mm$^{3}$ resolution allows us to process large sections of the body
at once. Using Tensorflow, we created a computationally-efficient
model which can be deployed on a wide variety of devices, for use
by practitioners of other disciplines \cite{tensorflow:2015}. 

Data augmentation has become standard practice in applying deep learning
models to images. As in our work, several others have applied random
affine transformations to 3D volumes \cite{Milletari:2016:Vnet}.
However, to our knowledge, we are the first to develop a computationally
efficient system for augmenting 3D images using GPU texture sampling,
and the first to leverage GPUs for random noise generation for the
purpose of data augmentation. Some of the current deep learning frameworks,
such as Keras, provide built-in operations for image manipulation
\cite{chollet:2015:keras}. However, at the time of writing these
do not support 3D geometric operations, and they are probably not
implemented in the most efficient way. 

Automatic generation of training labels is an active area of machine
learning research, sometimes called ``weak supervision.'' Ghafoorian
et al.~used a region-growing algorithm to train a deep neural network
for brain ventricle segmentation \cite{Ghafoorian:2018:StudentBeatsTeacher}.
We apply an even simpler method to the lungs and bone, which is accelerated
via 3D Fourier transforms, and unlike the previous work, requires
no user input. This is essential for bones, which are too numerous
to annotate manually.

The final contribution of our work is the IOU loss, both its proposal
and mathematical analysis. This is closely related to the Dice loss
which was proposed by Milltari et.~al \cite{Milletari:2016:Vnet}.
Sudre et al.~noted that the Dice loss, as well as a variant based
on fuzzy set theory, outperform weighted cross-entropy for unbalanced
classes \cite{Sudre:2017:GeneralizedDiceLoss}. However, to our knowledge,
no one has yet expounded on the mathematical properties of these loss
functions, their relationship to the binary scores for which they
are named, nor the reason for their superior performance to the usual
per-voxel loss functions \cite{Sudre:2017:GeneralizedDiceLoss}.

\section{Training label generation\label{sec:Training-label-generation}}

This section explains how we used image morphology to automatically
detect and segment organs, which generates training labels for our
neural network. First we describe the basics of $n$-dimensional image
morphology, and how we accelerated these operations using Fourier
transforms. Then we describe the specific algorithms used to segment
the lungs and bones.

\subsection{Morphology basics, acceleration by Fourier transforms}

Let $f:\mathbb{Z}^{n}\rightarrow\mathbb{F}^{2}$ denote a binary image,
and $k:\mathbb{Z}^{n}\rightarrow\mathbb{F}^{2}$ the structuring element.
Then dilation is 
\[
D(f,k)(x)=\begin{cases}
1, & (f\ast k)(x)>0\\
0, & \mbox{otherwise.}
\end{cases}
\]
That is, we first convolve $f$ with $k$, treating the two as real-valued
functions on $\mathbb{Z}^{n}$. Then, we convert back to a binary
image by setting zero-valued pixels to black, all others to white.
Erosion is computed similarly: if $\bar{f}$ is the binary complement
of $f$, then erosion is just $E(f,k)(x)=\overline{D(\bar{f},k)}$.
Similarly, the opening and closing operations are compositions of
erosion and dilation.

Written this way, all of the basic operations in $n$-dimensional
binary morphology reduce to a mixture of complements and convolutions,
the latter of which can be computed by fast Fourier transforms, due
to the identity $\widehat{f\ast k}=\widehat{f}\cdot\widehat{k}$,
where $\widehat{f}$ denotes the Fourier transform of $f$. This allows
us to quickly generate training labels by morphological operations,
especially when the structuring elements are large. In what follows,
we describe how morphology is used to generate labels for two organs
of interest, the skeleton and lungs. These organs were chosen because
their size and intensity contrast enable detection by simple thresholding.

\subsection{Morphological detection and segmentation of CT lungs\label{subsec:Morphological-detection-and}}

\begin{figure}
\begin{centering}
\includegraphics[scale=0.25]{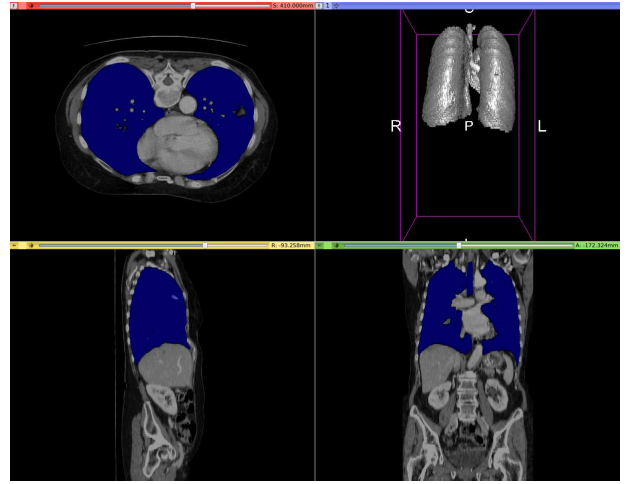}
\par\end{centering}
\caption{Example of CT lung detection and segmentation using image morphology.
Lung mask overlaid in blue. Rendered by 3D Slicer \cite{Kikinis:2014:SlicerOverview}.}

\label{fig:morph_lung}
\end{figure}

The lungs were detected and segmented based on the simple observation
that they are the two largest air pockets in the body. First, we extract
the air pockets from the CT scan by removing all voxels greater than
-150 Hounsfield units (HU). The resulting mask is called the thresholded
image. Then, we remove small air pockets by morphologically eroding
the image using a spherical structuring element with a diameter of
1 cm. Next, we remove any air pockets which are connected to the boundary
of any axial slice in the image\footnote{That is, the maximal and minimal extent of the image in the $x$ and
$y$ axes.}. This removes air outside of the body, while preserving the lungs.
From the remaining air pockets, we take the two largest connected
components, which are almost certainly the lungs. Finally, we undo
the effect of erosion by taking the components of the thresholded
image which are connected to the two detected lungs. An example output
in shown in figure \ref{fig:morph_lung}.

A quick web search reveals that similar algorithms have previously
been used to segment the lungs \cite{mathworks:2018:lungSeg}. The
novelty in our approach lies in using this algorithm to train a deep
neural network, which can be more robust than the morphological algorithm
from which it was trained, as discussed in section \ref{subsec:experiment_CT-organ-segmentation}.

\subsection{Morphological detection and segmentation of CT skeleton}

Bone segmentation proceeds similarly to lung segmentation, by a combination
of thresholding, morphology and selection of the largest connected
components. This time we define two intensity thresholds, $\tau_{1}=0$
and $\tau_{2}=200$ HU. These were selected so that almost all bone
tissue is greater than $\tau_{1}$, while the hard exterior of each
bone is usually greater than $\tau_{2}$. First, we select the exteriors
of all the bones by thresholding the image by $\tau_{2}$. This step
inevitably also includes some unwanted tissues, such as the aorta,
kidneys and intestines, especially in contrast-enhanced CTs. To remove
these unwanted tissues, we select only the largest connected component,
which is the skeleton. Next, we fill gaps in the exteriors of the
bones by morphological closing, using a spherical structuring element
with a diameter of 2.5 cm. This step has the unwanted side effect
of filling gaps between bones as well, so we apply the threshold $\tau_{1}$
to remove most of this unwanted tissue.

\begin{figure}
\begin{centering}
\includegraphics[scale=0.23]{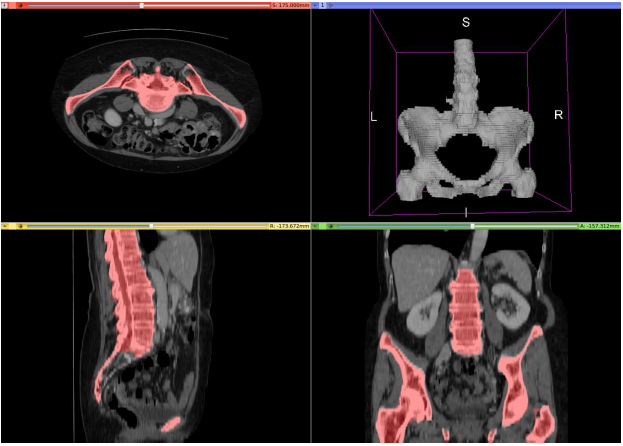}
\par\end{centering}
\caption{Example of CT skeleton detection and segmentation using image morphology.
Skeleton mask overlaid in red. Rendered by 3D Slicer \cite{Kikinis:2014:SlicerOverview}.}
\label{fig:3d_skeleton}
\end{figure}

At this stage, there could be holes in the center of large bones,
such as the pelvis and femurs. To fill these, we note that, when the
patient is reclined on the exam table, large bones almost always lie
parallel to the $z$-axis of the image. Accordingly, we process each
$xy$-plane (axial slice) in the image, filling in any holes which
are not connected to the boundaries. This fills in the centers of
large bones in most slices, which suffices for our purposes of training
a deep neural network. See figure \ref{fig:3d_skeleton} for a 3D
visualization of the resulting skeleton segmentation. The accuracy
of this scheme is evaluated in section \ref{subsec:experiment_CT-organ-segmentation}.

\section{GPU-accelerated data augmentation\label{sec:GPU-accelerated-data-augmentatio}}

Deep neural networks can be viewed as \textit{tabulae rasae} on which
a wide variety of classifiers can be inscribed, depending on the training
data. For example, we typically expect that image classifiers should
be invariant to rotation and scaling, but the basic units of convolutional
neural networks (CNNs) need not be. Given a basic dataset, data augmentation
applies random transformations to generate new training data. Theoretically,
this can be justified as modifying the data distribution, which alters
the expected loss for each choice of parameters. The expanded dataset
also helps to combat over-fitting.

Our organ segmenter was trained using a variety of data transformations
including affine warping, intensity windowing, additive noise and
occlusion. These common operations are expensive to compute over large
3D volumes. In particular, affine warping requires random access to
a large buffer of image data, with little reuse, which is inefficient
for the cache-heavy memory hierarchy of CPUs. A typical CT scan consists
of hundreds of $512\times512$ slices of 12-bit data. When arranged
into a 3D volume, a CT scan is hundreds of times larger than a typical
low-resolution photograph used in conventional computer vision applications. 

Although 3D data augmentation is slow on conventional CPUs, the operations
involved are actually very efficient when implemented on more specialized
hardware. All of the aforementioned operations are common in computer
graphics, so GPUs have been designed to handle them efficiently. In
particular, GPU texture memory is optimized for parallel access to
2D or 3D images, and includes special hardware for handing out-of-bounds
coordinates and interpolation between neighboring pixels. This is
especially well-suited to accelerating affine warping. Photometric
operations such as noise generation, windowing and cropping also benefit
from the massive parallelism offered by GPUs. 

\begin{figure}
\begin{centering}
\includegraphics[scale=0.45]{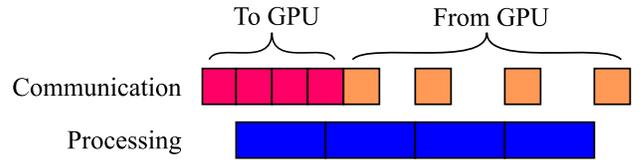}
\par\end{centering}
\caption{Pipelining jobs for efficient batch processing. With this scheme,
a batch of training images is augmented with minimal latency from
data transfers.}

\label{fig:memory_pipeline}
\end{figure}

Since these operations involve little reuse of data, each output pixel
is drawn by its own CUDA thread. Each thread computes the affine coordinate
map, samples the input image at that coordinate, applies the photometric
transformations, then writes the final intensity value to the output
volume. In this way, each output requires only a single access to
texture memory.. To mitigate the cost of transferring data between
memory systems, we designed our data augmentation library with a first-in
first-out (FIFO) queue to pipeline jobs. The concept is illustrated
in figure \ref{fig:memory_pipeline}. While one image is being processed,
the next has already begun transferring from main memory to graphics
memory, effectively hiding its transfer latency. The FIFO programming
model naturally matches the intended use case of augmenting an entire
training batch at once. Our experiments in section \ref{subsec:experiment_GPU-data-augmentation}
show this scheme accelerates processing by $1.69\times$.

In what follows, we describe our image manipulation pipeline, with
the details of each operation.

First, we warp the image using a random affine transformation. Sampling
coordinates are computed as $x^{\prime}=Ax+b$, where $x,b\in\mathbb{R}^{3}$
and $A\in\mathbb{R}^{3\times3}$. The matrix $A$ is generated by
composing a variety of geometric transformations drawn uniformly from
user-specified ranges, including rotation, scaling, shearing, reflection
and generic affine warping\footnote{For reflection, the user specifies the probability that the image
is reflected about each of the three axes.}. Finally, a random displacement $d\in\mathbb{R}^{3}$ is drawn from
a uniform distribution according to user-specified ranges. Taking
$c\in\mathbb{R}^{3}$ to be the coordinates of the center of the volume,
we then compute $b$ according to the formula $b=c+d-Ac$, which guarantees
$Ac+b=c+d$. That is, the center of the image is displaced by $d$
units. The output image is defined by $I_{\mbox{affine}}(x)=I_{\mbox{in}}(Ax+b)$,
where $I_{\mbox{in}}:\mathbb{R}^{3}\rightarrow\mathbb{R}$ denotes
the input image volume from the training set. The discrete image data
is sampled from texture memory using trilinear interpolation, whereas
the labels are sampled according to the nearest neighboring voxel.

Second, we apply random occlusion, also known as cropping. To encourage
robustness to missing anatomy, we randomly set part of the input volume
to zero, while ensuring that every voxel has an equal chance of being
occluded. The occluded region is a rectangular prism, with height
uniformly distributed in the interval$\delta\in[0,\delta_{\mbox{max}}]$,
and starting coordinate $z\in[-\delta_{\max},z_{\mbox{max}}]$, where
$z_{\mbox{max}}$ is the maximum possible $z$-coordinate. Then, we
compute
\[
I_{\mbox{occ}}(x)=\begin{cases}
0, & z\ge x_{3}\ge z+\delta\\
I_{\mbox{affine}}(x), & \mbox{otherwise}.
\end{cases}
\]
Since we are already applying an affine transformation to the image,
removing an axis-aligned prism from the output effectively removes
a randomly-oriented prism from the input. For efficiency, we evaluate
occlusion prior to sampling the image texture. If the value is negative,
all future operations are skipped, including the texture fetch.

Third, we introduce additive Gaussian noise as a simplistic model
of the artifacts introduced in image acquisition. The operation is
simply $I_{\mbox{noise}}(x)=I_{\mbox{occ}}(x)+n(x)$, where $n(x)$
is drawn from an independent, identically distributed (IID) Gaussian
process with zero mean and standard deviation $\sigma$. The sole
parameter $\sigma$ is drawn from a uniform distribution for each
training example. In this way, some images are severely corrupted
by noise, while others are hardly changed. We used the cuRand library
to quickly generate noise on the GPU \cite{curandWebsite}. We initialize
a separate random number generator (RNG) for each GPU thread, with
one thread per output voxel. To reduce the initialization overhead,
each thread uses a copy of the same RNG, starting at a different seed.
This sacrifices the guarantee of independence between RNGs, but the
effect is not perceivable in practice.

Finally, we apply a random window/level transformation to the image
intensities. To increase image contrast, radiologists always view
CT scans within a certain range of Hounsfield units (HU). For example,
bones might be viewed with a window of -1000-1500 HU, while abdominal
organs would be viewed with a narrower window of -150-230 HU. To
simulate this, we randomly draw limits $-\infty<a<b<\infty$ according
to user-specified ranges. Then we compute
\[
I_{\mbox{window}}(x)=\min\left\{ \max\left\{ \frac{I_{\mbox{noise}}(x)-a}{b-a},0\right\} ,1\right\} .
\]
In other words, the intensity values are clamped to the range $[a,b]$,
and then affinely mapped to $[0,1]$. This is straightforward to implement
on GPUs.

\section{Neural network architecture}

This section describes the design of our predictive model; both the
inputs and outputs, pre- and post-processing, the neural network itself,
and the training loss function.

\subsection{Pre- and post-processing}

Our neural network takes as input a $120\times120\times160$ image
volume, and outputs a $120\times120\times160\times6$ probability
map, where each voxel is assigned a class probability distribution.
This becomes a $120\times120\times160$ prediction map by taking the
$\arg\max$ probability for each voxel. To reduce memory requirements,
we resample all image volumes to $3$ mm$^{3}$ before they are fed
into the model. Resampling consists of Gaussian smoothing, which serves
as a lowpass filter to avoid aliasing artifacts, followed by interpolation
at the new resolution. Since each CT scan has its own millimeter resolution
for each dimension $u=(u_{1},u_{2},u_{3})$, we adjust the Gaussian
smoothing kernel according to the formula $g(x)\propto\exp\left(-\sum_{k=1}^{3}x_{k}^{2}/\sigma_{k}^{2}\right)$where
the smoothing factors are computed from the desired resolution $r=3$
according to $\sigma_{k}=\frac{1}{3}\max(r/u_{k}-1,0).$ This heuristic
formula is based on the fact from digital signal processing that,
in order to avoid aliasing, the cutoff frequency should be placed
at $r/u_{k}$, the ratio of sampling rates, on a $[0,1]$ frequency
scale.

After the neural network, we resample the $120\times120\times160$
prediction map to the original image resolution by nearest neighbor
interpolation. One difficulty with this scheme is that CT scans vary
in resolution and number of slices, and at $3$ mm$^{3}$ we are unlikely
to fit the whole scan in our network. For training, we address this
by selecting a $120\times120\times160$ sub-region from the scan uniformly
at random. For inference, we cover the scan by partially-overlapping
sub-regions, averaging predictions where overlap occurs. Others have
addressed the issue of limited viewing resolution by training multiple
FCNs, each working at different scales \cite{Roth:2018:CascadedFCNOrganSeg}.
However, we found that a single 3 mm$^{3}$ network achieves competitive
performance.

\subsection{Model}

We chose a relatively simple model which balances speed and memory
consumption with accuracy. Our model is based on GoogLeNet, but its
convolution and pooling operators work in 3D instead of 2D \cite{Szegedy:2015:GoogLeNet}.
Like other fully-connected networks (FCNs), our model essentially
consists of two parts: one for decimation and one for interpolation
\cite{Shelhamer:2017:FCN}. The decimation network is similar to the
usual convolutional neural network (CNN) used for image classification,
with max-pooling and strided convolution layers which reduce the image
size by a factor of 8 in each dimension. The interpolation layer restores
the feature maps to their image dimensions, essentially through convolutional
interpolation. Unlike most of the literature, we chose not to use
any ``skip connections'' which forward feature maps in the decimation
part to later layers in the interpolation part \cite{Shelhamer:2017:FCN,Milletari:2016:Vnet,Gibson:2018:DenseVnet}.
We did this two reasons: first, to save memory which is precious when
dealing with 3D models. Second, to disambiguate the advantages of
our loss function and data augmentation from any improvements in model
architecture.

\begin{table}

\caption{List of layers in our fully-convolutional neural network.}

\begin{centering}
\begin{tabular}{ccccc}
\hline 
Purpose & Type & Filter size & Outputs & Stride\tabularnewline
\hline 
\hline 
\multirow{13}{*}{Decimation} & Convolution & 7 & 64 & 2\tabularnewline
 & Convolution & 1 & 64 & 1\tabularnewline
 & Convolution & 4 & 192 & 1\tabularnewline
 & Pooling & 2 & 192 & 2\tabularnewline
 & Inception & 1-5 & 256 & 1\tabularnewline
 & Inception & 1-5 & 480 & 1\tabularnewline
 & Inception & 1-5 & 512 & 1\tabularnewline
 & Inception & 1-5 & 512 & 1\tabularnewline
 & Inception & 1-5 & 528 & 1\tabularnewline
 & Inception & 1-5 & 832 & 1\tabularnewline
 & Pooling & 2 & 832 & 2\tabularnewline
 & Inception & 1-5 & 832 & 1\tabularnewline
 & Inception & 1-5 & 1024 & 1\tabularnewline
\hline 
Interpolation & Interpolation & 8 & 6 & 1/8\tabularnewline
\hline 
Output & Softmax & n/a & 6 & 1\tabularnewline
\hline 
\end{tabular}
\par\end{centering}
\label{table:nn_layers}
\end{table}

Table \ref{table:nn_layers} lists the layers of our neural network,
in order from the input image data to the final probability maps.
The exact details of each layer type are beyond the scope of this
paper, but should be familiar to practitioners. Filter sizes and strides
apply to all three dimensions, for example a filter size of 7 implies
a $7\times7\times7$ isotropic filter. All convolutions are followed
by constant ``bias'' addition, batch normalization and rectification
\cite{Ioffe:2015:BatchNorm}. Pooling always refers to taking neighborhood
maxima. An inception module consists of a multitude of convolution
layers of sizes 1, 3 and 5, along with a pooling layer, which are
concatenated to form four heterogeneous output paths \cite{Szegedy:2015:GoogLeNet}.
The inception module seems to be a memory-efficient way to construct
very deep neural networks, since it employs relatively inexpensive
operations of heterogeneous sizes. For simplicity, we report the total
number of outputs of the inception module, rather than the number
of filters of each type. The final softmax layer outputs class probabilities
for each voxel.

\subsection{IOU Loss\label{subsec:IOU-Loss}}

The most common loss function used for image segmentation is weighted
cross-entropy. This method assigns a separate loss function to each
voxel, minimizing the weighted average of all losses. To compensate
for extreme class imbalances encountered in medical imaging, a separate
weight is assigned to each voxel so that all objects weigh the same,
regardless of size. We found this apporach to give poor results, and
instead formulated a different loss function, the IOU loss.

\begin{figure}
\begin{centering}
\includegraphics[scale=0.4]{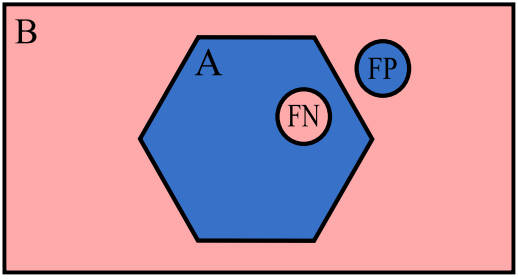}
\par\end{centering}
\caption{Example of segmentation errors with unbalanced classes. The blue color
represents a prediction of class A, the pink a prediction of class
B. The ground truth is the filled in pentagon, while the two circles
are prediction errors. With weighted cross-entropy, the pink circle
receives more weight than the blue circle, since the weight of each
pixel is determined by the size of the ground truth class.}
\label{fig:fp_weight}
\end{figure}

Imagine a binary classification scenario, as shown in figure \ref{fig:fp_weight}.
Let $p\in[0,1]^{n}$ denote the vector of $n$ output probabilities
from the model, where $n$ is the number of voxels, let $y\in\{0,1\}^{n}$
denote the binary ground truth labels, and let $w=(1/n)\sum_{k=1}^{n}y_{k}$
denote the weight of the class $y_{k}=1$. Let $L_{k}(p_{k})$ denote
the cross-entropy loss function for voxel $k$. Then weighted cross-entropy
loss is
\[
L_{\mbox{CE}}(p)=\frac{1}{n}\sum_{k=1}^{n}\left((1-y_{k})(1-w)+y_{k}w\right)L_{k}(p_{k}).
\]
The problem with this loss function is that it weights errors unequally:
a false positive receives the weight $1-w$, whereas a false negative
receives $w$. As a result, the model learns to exaggerate the boundaries
of small objects, since false positives almost always receive the
weight of the large ``background'' class. This is a limitation not
just of cross-entropy, but of any loss function which is a weighted
average over each voxel.

In order to address this limitation, we adopted a loss function which
minimizes the intersection over union (IOU) of the output segmentation,
also called the Jaccard index. Since the IOU depends on binary values,
it cannot be optimized by gradient descent, and thus it is not directly
suitable as a loss function for deep learning. Instead, our strategy
is to define a smooth function which is equal to the IOU when the
output probabilities are all either $0$ or $1$.

Since $\{0,1\}^{n}\subset[0,1]^{n}$, we can consider $y$ as a vector
in $[0,1]^{n}$ consisting only of probabilities $0$ and $1$. Then,
the smooth IOU loss is
\[
L_{\mbox{IOU}}=1-\frac{\|py\|_{1}}{\|p\|_{1}+\|y\|_{1}-\|py\|_{1}}
\]
where $\|p\|_{1}=\sum_{k}p_{k}$, since $p_{k}\ge0$ for all $k$.
To extend this to multi-class segmentation, simply average the $L_{\mbox{IOU}}$
for each class. This approach seems simpler and more naturally motivated
than the multi-class Dice scheme of Sudre et al.~\cite{Sudre:2017:GeneralizedDiceLoss}.

The IOU loss is closely related to the Dice loss, which was unceremoniously
proposed by Milletari et al.~\cite{Milletari:2016:Vnet}. The Dice
loss is

\[
L_{\mbox{Dice}}=1-\frac{2\|py\|_{1}}{\|p\|_{1}+\|y\|_{1}}.
\]

Since the IOU loss is similar to the Dice loss, one might wonder why
we bothered to propose a new loss function. One advantage is that,
when restricted to binary scores, the IOU loss obeys the triangle
inequality, and thus qualifies as a metric \cite{Spath:1981:JaccardMetric}.
This is not true of the Dice loss. For example, let $p=(0,1)$ and
$y=(1,0)$. Then it can be verified that $L_{\mbox{Dice}}(p,y)\ge L_{\mbox{Dice}}(p,p\cup y)+L_{\mbox{Dice}}(y,p\cup y)$,
which violates the triangle inequality.

One useful application of the triangle inequality is to bound the
amount by which we can decrease the IOU loss by restricting the experiment
to a subset of the domain. For example, for any set of binary predictions
$p$ and labels $y$, we can define a restriction by the taking the
intersection with a binary vector $s$. According to the triangle
inequality,
\begin{align}
L(p,g) & \le L(p,p\cap s)+L(g\cap s,p\cap s)+L(g,g\cap s),\label{eq:-2}
\end{align}
so the loss decrease is bounded by $L(p,p\cap s)+L(g,g\cap s)$, which
simplifies to $\|p\cap s\|_{1}/\|p\|_{1}+\|g\cap s\|_{1}/\|g\|_{1}$. 

To our knowledge, no one has yet described the mathematical properties
of the Dice loss in detail. Since it has essentially the same properties
as the IOU loss, both being continuous interpolations of binary or
set functions, we treat both losses simultaneously. They have the
following properties, which are easily verified:
\begin{enumerate}
\item They are equal to the desired binary loss, either Dice or IOU, when
$p$ is binary.
\item They are strictly increasing in each $p_{k}$ when $y_{k}=1$, decreasing
when $y_{k}=0$.
\item They are maximized only when $p=y$, minimized only when $p=1-y$.
\item They are smooth functions, if we define the loss to be $1$ at $p=y=0$,
which is otherwise undefined.
\end{enumerate}

Properties 1-3 ensure that minimizing the continuous loss corresponds
to maximizing the binary score of the trained model, while properties
2-4 encourage the training optimization problem to be well-behaved.
For example, property 2 implies that the function has no strict local
minima, since these would also be local minima when restricted to
a single input probability $p_{k}$. However, this property does not
extend to an average of per-image loss functions over a dataset, since
this could be a sum of both increasing and decreasing functions, which
need not be monotone.

Importantly, these properties do not suffice for uniqueness of the
loss function, since there are other possible behaviors when $p$
is nonbinary. In fact, these properties are not even unique among
the rational functions. For example, consider the family of loss functions
\[
L_{\mbox{IOU}}^{m}=\frac{\sum_{k=1}^{n}p_{k}^{m}y_{k}}{\sum_{k=1}^{n}p_{k}^{m}+\sum_{k=1}^{n}y_{k}-\sum_{k=1}^{n}p_{k}^{m}y_{k}}
\]
for any power $m>0$. It is easily verified that these functions also
satisfy all of the above properties, but differ from the $m=1$ case
for non-binary values of $p$. More generally, if $f_{1},\dots f_{n}$
is a collection of smooth increasing functions on $[0,1]$ with $f_{k}(0)=0$
and $f_{k}(1)=1$ for all $k\in\{1,\dots,n\}$, then the function
\[
L_{\mbox{IOU}}^{f}=\frac{\sum_{k=1}^{n}f_{k}(p_{k})y_{k}}{\sum_{k=1}^{n}f_{k}(p_{k})+\sum_{k=1}^{n}y_{k}-\sum_{k=1}^{n}f_{k}(p_{k})y_{k}}
\]
satisfies properties 1-4. This shows that there is nothing like a
unique choice of ``IOU loss'' or ``Dice loss,'' unless deeper
properties are discovered. However, we can at least say that the proposed
loss functions are the lowest-order rational functions satisfying
our properties, and thus the most computationally efficient.

We now analyze the penalty of each type of error, false positives
and false negatives, for the IOU loss. Let$\epsilon=\|p-y\|_{1}$
denote the misprediction error, and let $N=\|y\|_{1}$. A false negative
corresponds to $p_{k}\le y_{k}$ for all $k$, whence the loss is
\begin{align}
L_{FN} & =1-\frac{N-\epsilon}{(N-\epsilon)+N-(N-\epsilon)}\label{eq:}\\
 & =\frac{\epsilon}{N}.\nonumber 
\end{align}
Similarly, a false positive has $p_{k}\ge y_{k}$, which yields
\begin{align}
L_{FP} & =1-\frac{N}{(N+\epsilon)+N-N}\label{eq:-1}\\
 & =\frac{\epsilon}{N+\epsilon}.\nonumber 
\end{align}
So long as $N\gg\epsilon$, the two penalties are approximately equal.
This shows that, unlike weighted cross-entropy, the IOU loss strikes
a reasonable balance between each type of error. The assumption $N\gg\epsilon$
essentially means that the model is performing well on the training
data, which we should expect towards the end of training.

\section{Experimental Results}

This section describes our experiments validating the methods on real
data. First describe how we created our CT organ dataset. Then, we
measure the speed of our GPU data augmentation against a CPU baseline.
Finally, we train various organ segmentation networks, evaluating
their performance alongside the unsupervised label generators, first
on our own dataset, and then on a public challenge.

\subsection{Dataset\label{subsec:Dataset}}

In order to train and evaluate our organ segmentation system, we annotated
a set of 130 anonymized CT volumes exhibiting a wide variety of imaging
conditions, both with and without contrast, abdominal, thoracic and
full-body. The original images came from the Liver Tumor Segmentation
(LiTS) Challenge organized by Christ et al.~\cite{Christ:2017:LiTs}.
The LiTS challenge provides liver masks for almost all of the data,
which were extracted using a semi-automated segmentation method. To
complete the dataset, we added a few liver masks ourselves, and cropped
some of the image volumes to eliminate undesired artifacts. To this
we added our own masks for the kidneys and bladders. All of these
annotations were created using ITK-SNAP, a free tool for volumetric
image segmentation, using a mixture of active contours and manual
correction \cite{Yushkevich:2006:ITK-SNAP}.

Since training labels were automatically generated for the bones and
lungs, we must evaluate the model performance on a different dataset.
However, manually labeling these organs is extremely tedious, which
was the original motivation for developing the automatic labelers.
As a compromise, we manually labeled the lungs and bones in 10 out
of the 130 CT scans, again using ITK-SNAP. For the bones, we saved
time by starting with the automatic labels, and then manually correcting
the errors and omissions.

\subsection{Data augmentation speed\label{subsec:experiment_GPU-data-augmentation}}

After developing our GPU data augmentation program, we implemented
the same operations using SciPy, a popular library for scientific
computing \cite{scipy:2001}. We then compared the speed of the two
programs on our CT organ dataset. In order to maintain consistency
with our organ segmentation experiment, all images and labels were
resampled to a resolution of $3$ mm$^{3}$. To remove the effects
of file I/O, we wrote the whole dataset into main memory at the start
of each experiment. Execution times were averaged over 5 different
image batches. Our data augmentation code was written in C++ and CUDA
and controlled by a Python wrapper. All experiments were performed
on a single machine with an Intel Core i7-6900K CPU and four NVIDIA
GeForce Titan X Pascal GPUs.

\begin{figure}
\begin{centering}
\includegraphics[scale=0.51]{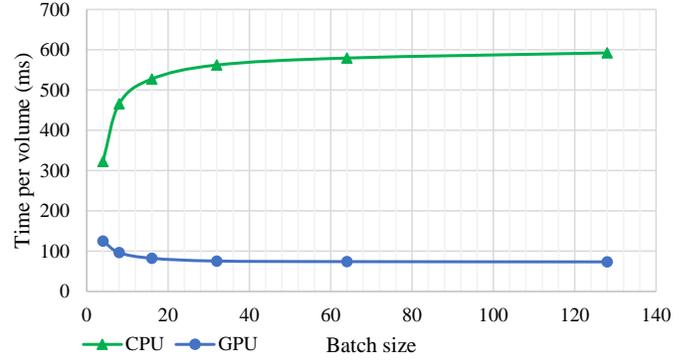}
\par\end{centering}
\caption{Comparison of our GPU data augmentation to a CPU implementation using
SciPy.}

\label{fig:benchmark}
\end{figure}

The average time per image volume is reported in figure \ref{fig:benchmark}.
The GPU implementation utilizes the memory transfer scheme from section
\ref{sec:GPU-accelerated-data-augmentatio}, so it benefits from increased
batch size which hides the communication overhead. Each batch is evenly
distributed across the four GPUs in our server, which allows for a
larger total batch size, albeit at marginal benefit over a single
GPU with a batch size of 32. On the other hand, the CPU grows slower
with increasing batch size, which is probably due to its multi-level
cache-based memory hierarchy. Execution times range from 125-74 ms
per CT scan on the GPU to 323-592 ms on the CPU. Accordingly, the
GPU offers a speedup of 2.6-8.1$\times$, depending on the batch size.
The largest GPU batch size is $1.69\times$ faster per volume than
the smallest, due to job pipelining. Comparing the fastest times for
each processor results in a $4.4\times$ speedup for the GPU\@. All
GPU times include the overhead of sending images to and from graphics
memory, which could theoretically be elided if deep learning frameworks
supported direct access to CUDA objects created by other programs.
This experiment shows that GPUs offer a considerable speedup over
generic CPUs for 3D data augmentation.

\subsection{CT organ segmentation accuracy\label{subsec:experiment_CT-organ-segmentation}}

\begin{table}
\caption{Average dice scores per CT volume for each organ segmenter.}

\label{table:organ_segmenation_accuracy}
\begin{centering}
\par\end{centering}
\begin{centering}
\begin{tabular}{cccccc}
\hline 
Method & \multicolumn{4}{c}{Neural Nework} & Morphology\tabularnewline
Loss & \multicolumn{2}{c}{Cross entropy} & \multicolumn{2}{c}{IOU} & n/a\tabularnewline
Data augmentation & No & Yes & No & Yes & n/a\tabularnewline
\hline 
\hline 
Lung & 87.5 & 86.8 & 90.8 & 91.8 & \textbf{97.8}\tabularnewline
Liver & 88.8 & 85.5 & 90.9 & \textbf{92.2} & n/a\tabularnewline
Bone & 73.6 & 65.4 & 78.9 & 79.3 & \textbf{93.2}\tabularnewline
Kidney & 71.8 & 65.2 & 77.5 & \textbf{78.3} & n/a\tabularnewline
Bladder & 70.4 & 58.4 & 80.1 & \textbf{83.7} & n/a\tabularnewline
Other & 97.7 & 96.4 & 98.6 & \textbf{98.7} & 99.7$^{*}$\tabularnewline
\hline 
\end{tabular}
\par\end{centering}
$^{*}$Includes all organs besides lung and bone
\end{table}

\begin{figure}
\begin{centering}
\includegraphics[scale=0.39]{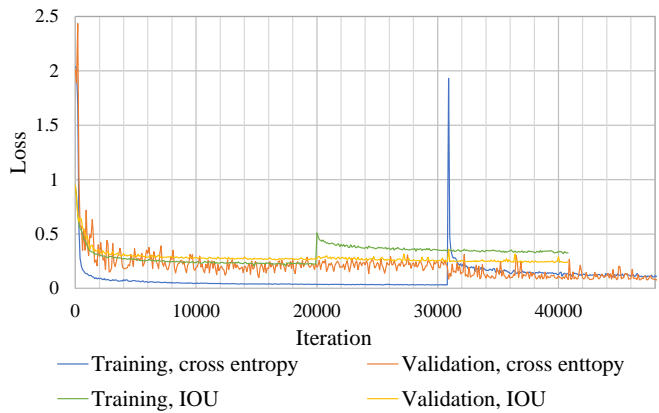}
\par\end{centering}
\caption{Learning curves for cross-entropy and IOU loss. The jump in training
loss corresponds to the introduction of data augmentation.}

\label{fig:learning_curves}
\end{figure}

\begin{figure}[t]
\begin{centering}
\includegraphics[scale=0.25]{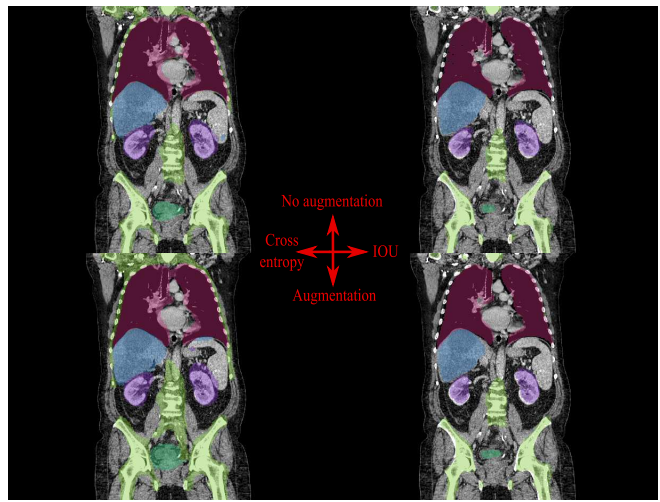}
\par\end{centering}
\caption{Results on a CT scan from the validation set, for each of the four
training schemes. IOU loss yields tighter boundaries than cross-entropy,
and IOU with augmentation is the only version yielding a reasonable
bladder.}

\label{fig:nn_output}
\end{figure}

In order to compare our proposed methods to the default variant, we
trained the same neural network with two different loss functions,
cross-entropy and IOU. In each case, we began without data augmentation,
only introducing it after the training loss ceased to decline. This
strategy saves time, since the network first learns the basic appearance
of each organ before adapting to the various augmentations. The learning
curves are shown in figure \ref{fig:learning_curves}. Each training
iteration consists of a batch of 4 image volumes. The training loss
is averaged over all the random crops in 100 training iterations,
while the validation loss is averaged over all sub-volumes of each
image in the validation set. In each case, training loss spikes upward
when augmentation is introduced, and then slowly settles back down.
Interestingly, data augmentation always increases the training loss,
as the training data becomes more difficult, but decreases the loss
on the unmodified validation data.

The final results are shown in table \ref{table:organ_segmenation_accuracy}.
For all organs, the IOU loss with data augmentation outperforms all
other variants. Even without augmentation, the IOU loss outperforms
both variants of cross-entropy. Surprisingly, data augmentation increases
validation accuracy with the IOU loss, but decreases the accuracy
with cross entropy. This occurs even though the validation cross-entropy
loss is decreasing, as shown in figure \ref{fig:learning_curves}.
This illustrates the issue described in section \ref{subsec:IOU-Loss},
where weighted cross-entropy encourages false positives over false
negatives for minority classes, so the model tends to overestimate
organ boundaries. This is especially evident with the bladder, as
shown in figure \ref{fig:nn_output}. The effect is less pronounced
for cross-entropy without data augmentation, since the model can over-fit
the small training set.

On our dataset, the neural network failed to match the accuracy of
the unsupervised morphological algorithm for bones and lungs. For
the largest and most distinctive organs, image morphology is simpler
and probably more precise than a neural network. However, the two
approaches are not mutually exclusive, as unsupervised methods can
post-process the output of a neural network. The main disadvantage
of morphology is that it is prone to catastrophic failure in ways
which are difficult to anticipate. For example, the morphological
lung segmenter can be thrown off by other air-filled objects in the
scan, such as an exam table. In contrast, a neural network captures
the visual appearance of each organ, which is often more reliable.
Finally, when segmenting a large variety of organs, a single network
has obvious conceptual and computational advantages over a litany
of brittle, organ-specific solutions. 

\subsection{Liver segmentation challenge}

Seeking external validation of our previous results, we ran the four
models on the LiTS challenge test dataset, without any additional
training for the liver-only task \cite{Christ:2017:LiTs}. Our results
are shown in table \ref{table:LiTS}. Our scores are similar to the
liver scores in table \ref{table:organ_segmenation_accuracy}, where
IOU loss with data augmentation outperforms all the other models. 

Our best model achieved a mean Dice per case of 90.5\%, which at the
time of writing would place us in rank 49 on that section of the challenge.
The liver segmentation leaderboard is highly competitive, as the top
score is 96.6\% out of a possible 100\%. Differences between such
high scores are likely attributable to subtle discrepancies in organ
boundaries, to which most applications are insensitive. Our model
is considerably simpler and more general than the top-scoring methods,
as it operates at a coarse 3mm$^{3}$ resolution, uses no additional
data, and identifies five different organs simultaneously. Our averge
processing time was 5.98s per volume, consuming a modest 2429 MB of
graphics memory, which is well within the capabilities of commodity
graphics cards. In contrast, other submissions focused only on the
liver, and leveraged significantly more complex and expensive methods,
including higher processing resolution, cascades of 2D and 3D models,
sophisticated post-processing, and even transfer learning from other
datasets \cite{Li:2018:H-DenseUNet_LiTS}. Our work shows that, with
the right training, a relatively simple model suffices for many applications.
Furthermore, nothing precludes the top-performing models from incorporating
our suggested improvements.

\begin{table}
\caption{Results of each liver segmenter on the LiTS challenge.}

\begin{centering}
\begin{tabular}{ccccc}
\hline 
Loss & \multicolumn{2}{c}{Cross entropy} & \multicolumn{2}{c}{IOU}\tabularnewline
Data augmentation & No & Yes & No & Yes\tabularnewline
\hline 
\hline 
Avg. Dice & 0.866 & 0.840 & 0.896 & \textbf{0.905}\tabularnewline
Global Dice & 0.872 & 0.846 & 0.904 & \textbf{0.911}\tabularnewline
VOE & 0.233 & 0.273 & 0.185 & \textbf{0.168}\tabularnewline
RVD & -0.193 & -0.248 & \textbf{-0.024} & 0.033\tabularnewline
ASSD & 8.506 & 10.407 & 5.876 & \textbf{3.767}\tabularnewline
MSSD & 0.767 & 173.722 & 88.843 & \textbf{54.832}\tabularnewline
RMSD & 18.299 & 21.908 & 13.533 & \textbf{8.380}\tabularnewline
\hline 
\end{tabular}
\par\end{centering}
\label{table:LiTS}

\end{table}

\section{Conclusion}

We delineated a dataset of 130 abdominal CT scans using a mixture
of manual annotation and automated morphological segmentation. We
used this dataset to train a deep neural network for CT organ segmentation,
using GPUs to accelerate data augmentation. Our model uses the IOU
loss to improve segmentation accuracy. We explained mathematically
why there is no unique IOU loss, and why the various IOU loss functions
outperform weighted cross-entropy for unbalanced segmentation tasks.
The code, data and trained model will be made publicly available.
We hope that the dataset will enable the development and evaluation
of more accurate organ segmentation methods. We invite others to annotate
new organs, which could easily be incorporated into the existing system.

\subsection*{Acknowledgments}

This work was supported in part by grants from the National Cancer
Institute, National Institutes of Health, 1U01CA190214 and 1U01CA187947.

\bibliographystyle{ieeetr}
\bibliography{organSeg18}

\end{document}